\documentclass[letterpaper, 10 pt, conference]{ieeeconf}  % Comment this line out if you need a4paper

\IEEEoverridecommandlockouts                              % This command is only needed if 
\usepackage{hyperref}
\hypersetup{hidelinks,
	colorlinks=true,
	allcolors=black,
	pdfstartview=Fit,
	breaklinks=true}
\usepackage{graphics} % for pdf, bitmapped graphics files
\usepackage{epsfig} % for postscript graphics files
\usepackage{mathptmx} % assumes new font selection scheme installed
\usepackage{times} % assumes new font selection scheme installed
\usepackage{amsmath} % assumes amsmath package installed
\usepackage{amssymb}  % assumes amsmath package installed
\usepackage{algorithm}
\usepackage{algorithmic}
\title{\LARGE \bf
WeatherDepth: Curriculum Contrastive Learning for Self-Supervised Depth Estimation under Adverse Weather Conditions
}

\author{Jiyuan Wang, Chunyu Lin$^{\dagger}$, Lang Nie, Shujun Huang, Yao Zhao, Xing Pan and Rui Ai
\thanks{Institute of Information Science, Beijing Jiaotong University, Beijing 100044, China. Haomo AI Technology Co.,Ltd, Beijing 100084, China. Email: \{wangjiyuan, cylin,nielang, shujuanhuang, yzhao\}@bjtu.edu.cn,\{panxing,luxiuyuan\}@haomo.ai.
}
\thanks{$^{\dagger}$ Corresponding author.
}
}
\usepackage{graphicx}
\usepackage{amsmath}
\usepackage{ulem}
\begin{document}
\maketitle
\thispagestyle{empty}
\pagestyle{empty}
\begin{abstract}
Depth estimation models have shown promising performance on clear scenes but fail to generalize to adverse weather conditions due to illumination variations, weather particles, etc. In this paper, we propose WeatherDepth, a self-supervised robust depth estimation model with curriculum contrastive learning, to tackle performance degradation in complex weather conditions. Concretely, we first present a progressive curriculum learning scheme with three simple-to-complex curricula to gradually adapt the model from clear to relative adverse, and then to adverse weather scenes. It encourages the model to gradually grasp beneficial depth cues against the weather effect, yielding smoother and better domain adaption. Meanwhile, to prevent the model from forgetting previous curricula, we integrate contrastive learning into different curricula. By drawing reference knowledge from the previous course, our strategy establishes a depth consistency constraint between different courses toward robust depth estimation in diverse weather. Besides, to reduce manual intervention and better adapt to different models, we designed an adaptive curriculum scheduler to automatically search for the best timing for course switching. In the experiment, the proposed solution is proven to be easily incorporated into various architectures and demonstrates state-of-the-art (SoTA) performance on both synthetic and real weather datasets. Source code and data are available at \url{https://github.com/wangjiyuan9/WeatherDepth}.
\end{abstract}

\section{INTRODUCTION}
\label{section19}
Depth estimation builds a bridge between 2D images and 3D scenes and has numerous potential applications such as 3D reconstruction \cite{r1}, autonomous driving, etc. 
In recent years, due to the high costs of GT-depth collection from LiDARs and other sensors, researchers have turned to self-supervised solutions by exploiting photometric consistency between the depth-based reconstructed images and the target images.
However, there is a sharp drop in depth precision when it comes to adverse weather conditions because weather particles spoil the consistency assumption and illumination variations produce an inevitable domain gap.
Recent works tried to mitigate the performance degradation by restoring clear weather scenes \cite{r22}, extracting features consistent with sunny conditions \cite{r3,r5}, knowledge distillation from clear scenes \cite{iccv,iccv2}, etc. However, these solutions do not account for the fact that weather comes in varying degrees and categories, and their data augmentation cannot reflect real situations well (Fig. \ref{Fig2}), which hinders the potential of the estimation algorithm under weather conditions.

In this paper, we propose a self-supervised robust depth estimation model (named WeatherDepth) to address the above issues through curriculum contrastive learning. On the one hand, we simulate the progressive advances from clear to relatively adverse, and then adverse weather scenes, building three simple-to-complex curricula with adverse weather to different degrees. Concretely, we first train a base model on sunny data with clear structures to satisfy photometric consistency. This allows the model to obtain better-generalizable local optima for pre-training on more complex scenarios \cite{r24}. Then we optimize the model on relative adverse weather images with light effects, ground snow and water, which share a part of common regions with the clear domain and inspire the model to gradually grasp the depth cues against the missing textures and contrasts. Finally, we train the model on adverse data with the addition of weather particles (e.g., raindrops), further boosting the capability of handling complex noise patterns and violations of self-supervised assumptions.

\begin{figure}
\centering
\includegraphics[width=\columnwidth]{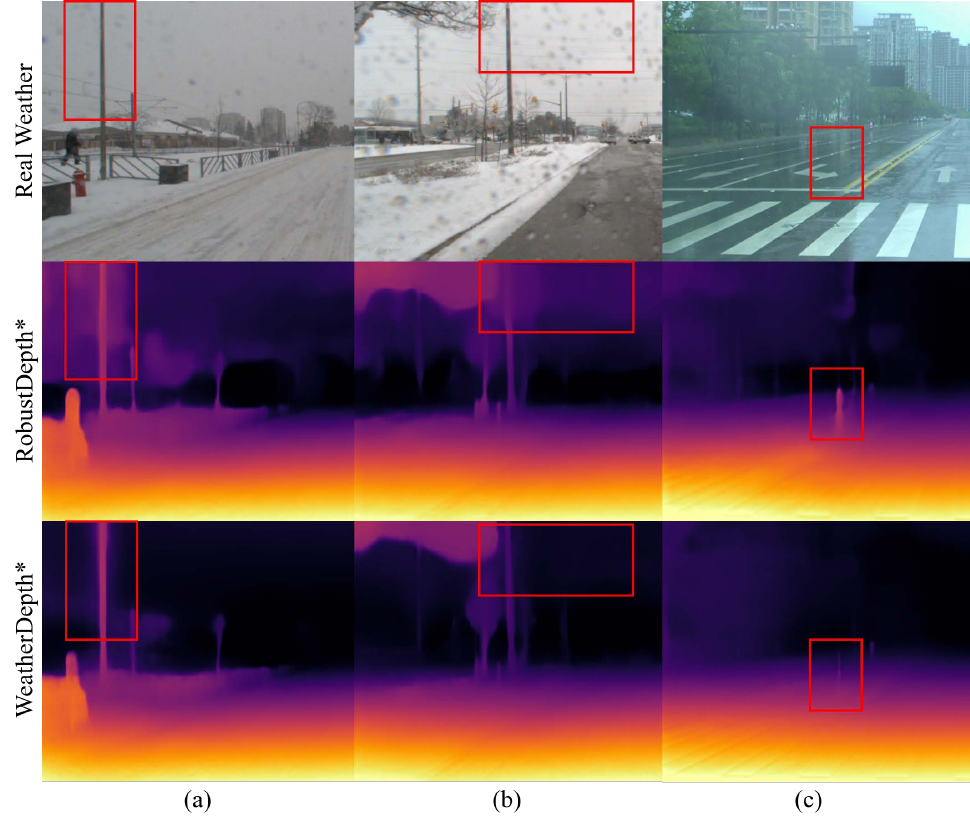}
\vspace{-0.6cm}
\caption{\textbf{Typical examples on real weather images.} Compared with Robust-Depth* (the SoTA robust depth estimation model under adverse weather), our WeatherDepth* produces more accurate results against  (a) snowflakes,  (b) raindrops on the lens, and (c) water surface reflections. Note both solutions adopt the same baseline model (MonoViT).
}
\label{Fig1}\vspace{-0.6cm}
\end{figure}

\begin{figure}
\centering
\includegraphics[width=\columnwidth]{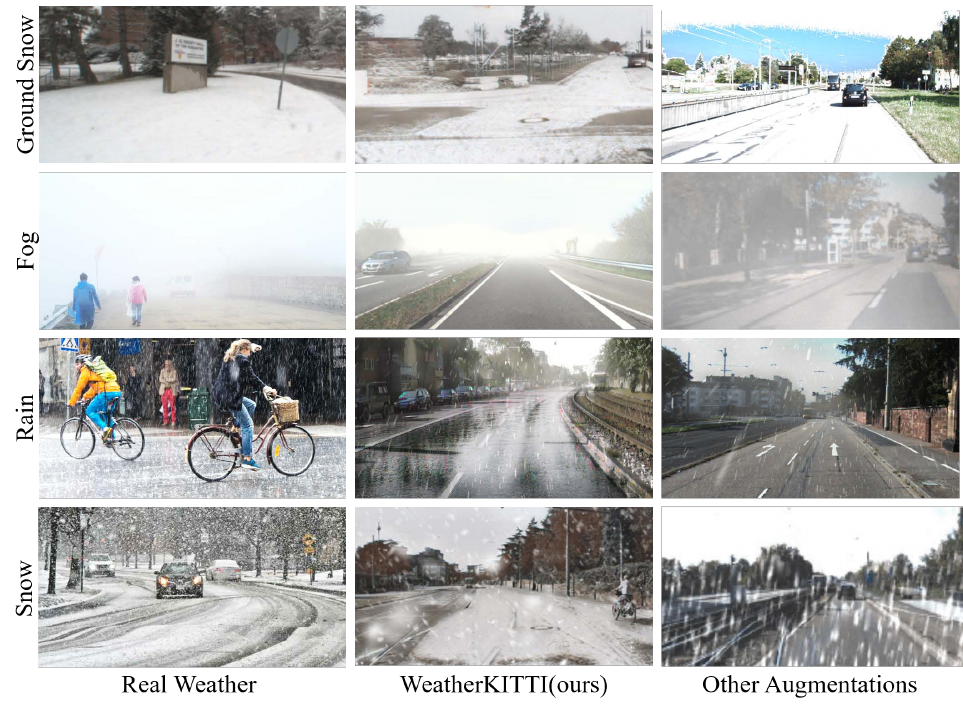}
\vspace{-0.6cm}
\caption{\textbf{Comparison of simulated adverse weather.} The other augmentations in the third column are from previous weather depth estimation studies \cite{iccv,r22,rbt}, which also adopt data augmentation. Obviously, our WeatherKITTI augmentation is significantly more natural than their results.
}
\label{Fig2}\vspace{-0.6cm}
\end{figure}

On the other hand, pre-defined curriculum learning alone may lead to catastrophic forgetting \cite{r9} due to the substantial inter-domain difference in each stage. To this end, we embed light-weight contrastive learning designs in different curricula. Specifically, as shown in Fig. \ref{Fig3}, we first establish \textbf{one} contrastive mode between two clear images with different traditional enhancements\cite{r13}. This forces the network to become more robust to this depth-irrelevant information and get prepared for the weather variation in later courses. Then, we build  \textbf{three} more challenging contrastive modes between the sunny scene and randomly selected rainy/snowy/foggy weather scenes. It effectively prevents the network from solely focusing on resisting weather changes and completely biasing its domain to the new weather.
In the last curriculum, we contrast three adverse weather against three relative adverse weather, constructing \textbf{nine} contrastive modes to improve the cross-weather robustness and relieve the problem of forgetting. 
These increasingly challenging contrastive modes(ranging from 1 to 3 to 9) formulate another curriculum learning process based on contrastive difficulty, which guides the training easily to be converged.

Moreover, we propose an adaptive curriculum scheduler to automatically switch curricula. It reduces manual intervention and produces smoother course transitions. To train an expected model and shrink the domain gap to real weather conditions, we combine GAN and PBR techniques \cite{r8} to build the WeatherKITTI dataset with diverse categories and magnitudes of weather. Compared with existing augmented weather data \cite{iccv,rbt,r22}, it renders a more realistic weather scene, as shown in Fig. \ref{Fig2}.

Finally, we incorporate the proposed curriculum contrastive learning scheme into three popular depth estimation baselines (PlaneDepth, WaveletMonodepth, and MonoViT) to evaluate its effectiveness. Experimental results show the proposed WeatherDepth models outperform the existing SoTA solutions on both synthetic and real weather datasets. To our knowledge, this is the first work to apply curriculum contrastive learning to depth estimation. To sum up, the main contributions are summarized as follows:
\begin{itemize}

    \item  To adapt to adverse weather without knowledge forgetfulness, we propose a curriculum contrastive learning strategy with robust weather adaption. It can be applied to various SoTA depth estimation schemes and could be extended to other simple-to-complex prediction tasks.

    \item To reduce manual intervention and better adapt to different models, an adaptive curriculum scheduler is designed to automatically switch the course by searching for the best timing. Besides, we built the WeatherKITTI dataset to narrow the domain gap to real weather situations.

    \item We conduct extensive experiments to prove the universality of our curriculum contrastive learning scheme on various architectures and the superior performance over the existing SoTA solutions.

\end{itemize}

\section{Related work}
\subsection{Self-supervised Depth Estimation}
Since the pioneering work of Zhou et al. \cite{zhou} showed that only using geometric constraints between consecutive frames can achieve excellent performance, researchers have continued to explore the cues and methods to train self-supervised models through videos \cite{r13,r15,r3} or stereo image pairs \cite{wav,r17,mdp1}. Afterward, methods including data augmentation\cite{r14}, self-distillation\cite{pld,r19}, indoor scenes aiding\cite{r20}, etc. have been introduced to self-supervised models, pushing their inference performance closer to supervised models. 
Our model adopts both supervised training manners, monocular and stereo, to verify the scalability of our method.

\subsection{Adverse Condition Depth Estimation}
Recently, the progress in depth prediction for typical scenes has opened up opportunities for tackling estimation in more challenging environments. Liu et al. \cite{r5} boost the nighttime monocular depth estimation (MDE) performance by using a Generative Adversarial Network (GAN) to render nighttime scenes and leveraging the pseudo-labels from day-time estimation to supervise the nighttime training. Then, Zhao et al. \cite{r3} consider rainy nighttime additionally. In this work, to fully extract features from both scenes, they used two encoders trained separately on night and day image pairs and applied the consistency constraints at the feature and depth domains. The first MDE model under weather conditions was proposed in \cite{iccv}. This work introduced a semi-augmented warp, which exploits the consistency between the clear frames while using the augmented prediction. Moreover, bi-directional contrast was incorporated in this work to improve the accuracy, although this doubles the training time. In \cite{iccv2}, instead of using the KITTI dataset that only contains dry and sunny scenes, NuScenes and Oxford RobotCar datasets were adopted, for their real rainy and night scenarios. They first train a baseline on sunny scenes, then fix these net weights and transfer-train another network for weather scenes with day distill loss.

Besides the above methods that combine data augmentation and various strategies, there are also other solutions \cite{r22} trying to estimate depth after removing the weather influence on the image. Our approach synthesizes the strengths of previous works, utilizing a single end-to-end encoder-decoder network architecture to build an efficient and effective solution.

\begin{figure*}
\centering
\includegraphics[width=1.0\textwidth]{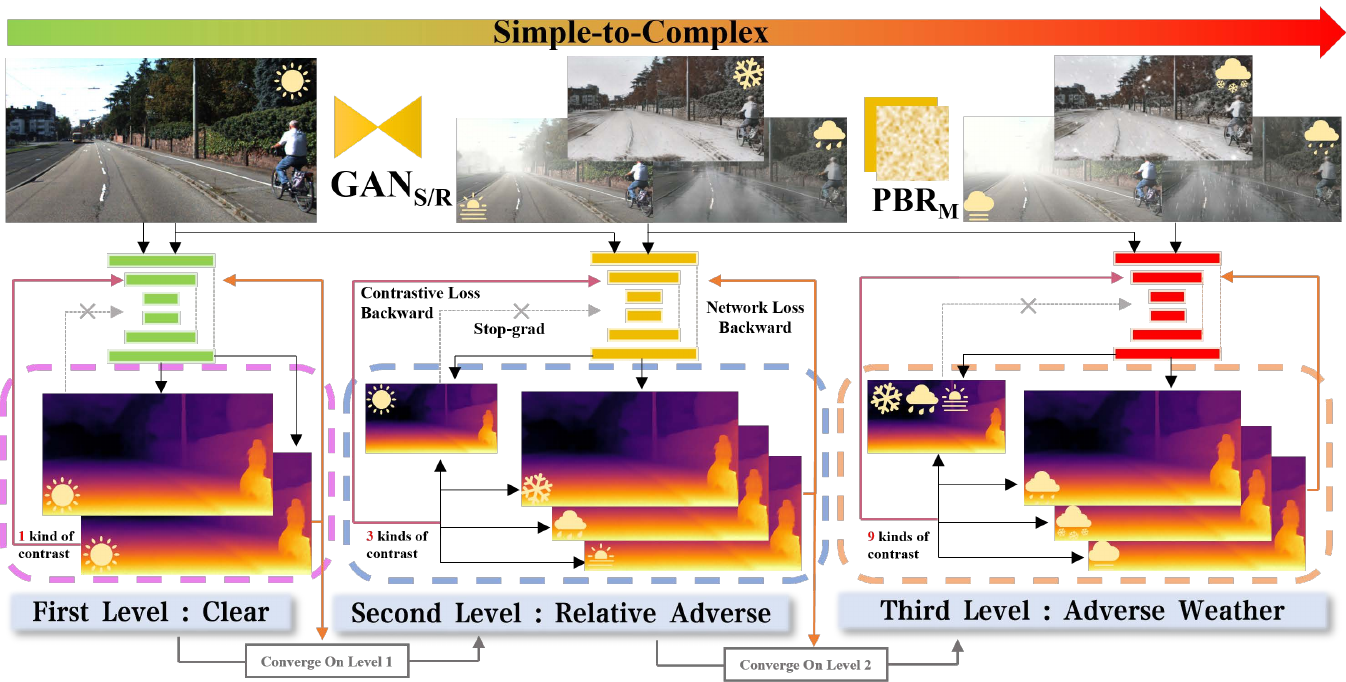}
\vspace{-0.45cm}
\caption{\label{Fig3}
\textbf{WeatherDepth pipeline} Through three progressive stages, our model-agnostic approach can estimate depth reliably under weather environments. Except for the last stage, we input the loss of estimation model into the curriculum scheduler, to change the level properly. And we input image pairs $I_{aug}$ and $I_{cst}$ to obtain depth maps $D_{aug}$ and $D_{cst}$. $D_{cst}$ is detached as the contrastive target to compute contrastive loss, which is weighted and backpropagated together with the original loss.}
\vspace{-0.45cm}
\end{figure*}

%%%%%%%%%%%%%%%%%%%%%%%%%%%%%%%%%%%%%%%%%%%%%%%%%%%%%%%%%%%%%
\vspace{-0.3cm}
\section{Method}
\vspace{-0.1cm}
\label{section18}
In this section, we elaborate on the key components and algorithms of the proposed method. 

\subsection{Preliminary}
\vspace{-0.1cm}
\label{section20}
The proposed WeatherDepth is built on self-supervised depth estimation. Given a target input image $I\in \mathbb{R}^{C \times H \times W}$ and an auxiliary reference image $I'$ from the stereo pair or adjacent frames, the self-supervised model $\mathcal{F}: I \rightarrow d \in \mathbb{R}^{H \times W}$ is expected to predict the disparity map $d$. 
With known baseline $b$ and focal length $f$, we compute the depth map $D=bf/d$. Then we can warp $I'$ to the target view using the projected coordinates that are generated from $D$, relative camera poses $T_{I' \rightarrow I}$ (from pose network or extrinsic) and intrinsic $K$:
\begin{equation}
\hat{I'} = I'\langle \text{Proj}(D, T_{I' \rightarrow I}, K) \rangle.
\end{equation}
The above equation describes such a warping process, in which $\hat{I'}$ denotes the warped image. The photometric reconstruction loss is then defined as: 
\begin{equation}
l_{ph}(d) = \alpha \frac{1-\text{SSIM}(I, \hat{I'})}{2} + \beta |I - \hat{I'}|.
\end{equation}
Based on stereo geometry, $l_{ph}$ equals 0 when $d$ is perfectly predicted. By minimizing the above loss, we can obtain the desired depth estimation. In addition, our WeatherDepth also adopts the semi-augmented warping from\cite{iccv}:
\begin{equation}
\tilde{I'}=I'\left\langle\operatorname{Proj}\left(\tilde{D}, T_{I' \rightarrow I}, K\right)\right\rangle,
\end{equation}
where $\tilde{D}$ is the depth estimated using augmented images, $\tilde{I'}$ is our semi-augmented warp result that will replace $\hat{I'}$ to calculate $l_{ph}$. We leverage the consistency between unaugmented images and the estimated depth of augmented images, avoiding the inconsistency between $\hat{I'_{aug}}$ and $I$ caused by weather variations.

\subsection{Curriculum Selection}
Due to the diversity of lighting and noise patterns in adverse weather, training directly on adverse weather data can easily lead to underfitting. To address this issue, we designed three curricula to gradually adapt to the new data domain. In curriculum designs, we obey the two principles: (1) The curriculum scenarios should follow the real-world weak-to-strong weather variation. (2) The curricula should be organized in a simple-to-complex order. 
To this end, we define our first curriculum as sunny scenes with slight adjustments on brightness, contrast, and saturation to make our model robust to these depth-invariant conditions.
Then we simulate the relative adverse weather by incorporating the groundwater reflections, ground snow, and droplets on the lens in the second course, which are not fully considered by previous works\cite{iccv2,iccv}. These effects not only create wrong depth cues (like Fig. \ref{Fig1} (b,c)) but also change the texture of the origin scenes.
In the last stage, we further introduce the raindrops, fog-like rain \cite{r8}, the veiling effect, and snow streaks \cite{r27}, because these particles are only visible in extremely adverse weather.

\vspace{-0.15cm}
\subsection{Curriculum Contrastive Learning}
\label{section21}
\vspace{-0.1cm}
To prevent the problem of forgetting the previous curricula, we embed contrastive learning into our curriculum learning process in an efficient manner. 

As depicted in line 6 of Algorithm \ref{alg1}, in ContrastStep, we use the TrainStep model to directly infer the depth of $I_{cst}$. Here $I_{clr}$, $I_{cst}$ and $I_{aug}$ should have the same depth since they are different weather augmentations of the same image, and weather changes do not affect the scene itself. Moreover, prediction will be more accurate in previous levels, for weather variations are less severe. Therefore, as shown in Fig. \ref{Fig3}, we can contrast the depth results from different curriculum stages to obtain the contrastive loss:
\begin{equation}
L_{\text{cst}} =
\begin{cases}
\log(|D_{aug} - \underline{D_{cst}}| + 1), & \text{if } S_{aug} > S_{cst} \\
\log(|\underline{D_{aug}} - D_{cst}| + 1), & \text{if } S_{cst} > S_{aug},
\end{cases}
\end{equation}
where $S_{aug}$ is the current curriculum stage, $S_{cst}$ is the stage of the contrastive weather. $D_{aug}$ and $D_{cst}$ are the depth predictions of the training image and contrastive inference, respectively. The underlining signifies that the gradients are cut off during backpropagation. Then the final loss is:
\begin{equation}
L_{\text{backward}} = L_{\text{model}} + w_{curr}\cdot L_{\text{cst}},
\end{equation}
where $L_{\text{model}}$ is the self-supervised model loss, $w_{curr}$ is the contrastive weight. Considering the model needs to adapt to weather changes when entering a new stage, we initialize it to a small value and update $w_{curr}$ each epoch according to:
\begin{equation}
w_{curr}=
\begin{cases}
w_{cst}, &\text{if}\ r=0 \ \\
\max(w_{max} w_{cst}, \lambda w_{curr}), & \text{if } r\neq0 \ \text{and} \ r\mid 2\\
w_{curr}, & \text{others},
\end{cases}
\end{equation}
where $\lambda$ is a constant $>1$. As the model gradually adapts to the curriculum stage, we want to steadily increase the consistency constraint. $r$ is the number of epochs trained in the current stage.

With the integration of contrast, our curriculum learning has considered both weather changes in curricula and contrastive difficulty alterations.

\subsection{Adaptive Curriculum Scheduler}
\label{section22}
\vspace{-0.1cm}
As mentioned in \cite{r9}, curriculum learning paradigms typically consist of two key components: the Difficulty Measurer and the Training Scheduler. The former is used to assign a learning priority to each data/task, while the latter decides when to introduce hard data into training. In this work, our Difficulty Measurer is pre-defined, whose benefits have been elaborated in section \ref{section19}. However, a pre-defined switching mode for a certain model may not adapt well to other networks. 
To this end, we check whether the network has fitted well in the current stage based on the change of self-supervised loss($L_{model}$). Concretely, we assume $L_{model}$ should always be decent when the network is not converged. So as shown in lines 10-20 in Algorithm \ref{alg1}, we record and average the $L_{model}$, and set a patience for every stage. We move to the next stage when the $L_{model}$ difference between the current and previous epoch surpasses the patience threshold. The contrastive loss is not included, because the weight of contrastive learning itself varies across epochs, adding it would make the model extremely unstable. This strategy is inspired by early stopping methods\cite{r28}, which effectively reduces training time and avoids overfitting at a certain stage.

\begin{algorithm}[tb]
\caption{Curriculum Scheduler for WeatherDepth}
\label{alg1}
\textbf{Input}: Clear data $I^{clr}_L,I^{clr}_R$, augmented data $I_i^{aug}$, contrast data $I_i^{cst}$, curriculum patience $P_i$($i = 1,2,3$ indicating the augmentation magnitude)
\begin{algorithmic}[1]
\STATE Let level $l=1$, patience $p=0$
\FOR{each epoch}
\STATE Update contrastive weight $w_{curr}$
\FOR{each batch}
\STATE $D^{aug},L_{model}$=TrainStep$(I_l^{aug},I^{clr}_L,I^{clr}_R)$
\STATE $D^{cst}$=InferenceStep$(I_l^{cst})$
\STATE $L_{cst}$=ContrastStep$(D^{aug},D^{cst}) $
\STATE recordloss.Append($L_{model}$)
\ENDFOR
\STATE recordkey.Append(Average(recordloss))
\IF {recordkey[-1] - recordkey [-2] $>$ threshold}
\STATE $p = p+1$
\ENDIF
\IF{$p \geq P_i$}
\STATE Reset $w_{curr}$ and $p = 0$
\STATE Switch to next-level data
\IF{$l$ = maxlevel and $P_i \geq 3$}
\STATE Load best epoch for the level $l$
\ENDIF
\STATE  $l=l+1$
\ENDIF
\ENDFOR
\end{algorithmic}
\end{algorithm}

\subsection{Method Scalability}
\label{section23}

To prove our expansiveness, we choose three popular models with tremendous differences in architecture as our baselines. The reasons are summarised as follows: 

\begin{itemize}
    \item \textbf{WaveletMonodepth}\cite{wav}: In this solution, wavelet transform is taken as the depth decoder, which trades off depth accuracy with speed. This is well suited for scenarios with different accuracy requirements.
    \item \textbf{PlaneDepth}\cite{pld}: This model uses Laplace mixture models based on orthogonal planes to estimate depth. It predicts depth for each plane respectively and computes the final depth by summing it over the Laplace distribution probabilities. Besides, PlaneDepth achieves current \textbf{SoTA} performance on self-supervised depth estimation.
    \item \textbf{MonoViT}\cite{vit}: Different from convolutional networks, MonoViT takes the transformer model as an encoder to improve image feature extraction. It represents the category of transformer-based models.
\end{itemize}

\def\tablename{\large Table}
\section{Experiments}
\label{section4}

\subsection{Dataset}
\label{section30}
\textbf{WeatherKITTI Dataset}: Based on KITTI \cite{kitti}, we establish a weather-augmented dataset to enhance depth estimation models with generalization to real weather. It contains three weather types that significantly impact visual perception: rain, snow, and fog. As shown in Fig. \ref{Fig3}, each weather type has two levels of magnitudes: relative adverse and adverse. The former includes light effects, ground snow and water, which are rendered by a CycleGAN model \cite{cyclegan} trained on CADC \cite{cadc}, ACDC\cite{acdc} and NuScenes\cite{nuscenes} datasets. The latter further adds noticeable weather particles through physically-based rendering. Following SoTA weather rendering pipelines \cite{r8, r27}, we generate adverse rains and snow masks. 
For foggy conditions, based on the atmospheric scattering model, we construct the second and third-stage foggy simulation augmentations under 150m and 75m visibility, respectively. Our rendered dataset covers all the images in the training and test scenes, totaling 284,304 (47,384$\times$6) images. To benchmark the robustness of models, we combine the original KITTI dataset with our weather-augmented KITTI dataset, naming it the WeatherKITTI dataset.

\textbf{DrivingStereo}\cite{stereo}: To characterize the performance of our models in real-world conditions, we use this dataset that provides 500 real rainy and foggy data images.

\textbf{CADC}: \cite{cadc} is one of few snowy datasets. However, its data is in sequential order. Therefore, we sampled 1 in every 3 sequential images and obtained 510 images as test data. As shown in Fig. \ref{Fig1}(a,b), this dataset contains real-world scenes with lens droplets, heavy snowfall, ground snow, etc. For depth GT generation, since LiDARs can be inaccurate under snowy conditions \cite{r2}, we utilize the DROR\cite{dror} algorithm to filter out erroneous depths caused by snowflakes and generate the final depth GTs. In addition, invalid sky and ego-vehicle regions(without Lidar points) are removed, with a final image resolution of 1280$\times$540 pixels.

\subsection{Implement Details}
\label{section31}
WaveletMonodepth, PlaneDepth, and MonoViT are three kinds of typical depth estimation models, and their performances are currently the best. Hence, we adapt the proposed curriculum contrastive learning scheme into these three baselines to validate our generalization, named as WeatherDepth, WeatherDepth$^*$, WeatherDepth$^{\dagger}$ in Table \ref{table1}. Most hyperparameters(learning rate, optimizer, training image resolution, etc.) are the same as their original implementation\cite{vit}\cite{pld}\cite{wav}. All models are trained on the WeatherKITTI dataset automatically with our scheduler. For stereo training, we use Eigen split \cite{eigen}. As for monocular training, we follow Zhou split \cite{zhou} with static scenes being removed. In our contrastive learning, we set $P_1=P_2=1$ and $w_{max}=10$ (all symbols are the same with Algorithm \ref{alg1}) for the three models. The model-specific minor modifications are shown in Table \ref{table1}. For all models, we strictly follow the original setup of \cite{vit, pld, wav} to train and evaluate our models.  In particular, for MonoViT, we use monocular training for 30 epochs and test on $640\times 192$ resolution. The contrast depth is not detached during MonoViT training. For PlaneDepth, we only adopt the first training stage \cite{pld}, training for 60 epochs in a stereo manner and testing with $1280 \times384$ resolution. For WaveletMonodepth, we train the Resnet-50 version for 30 epochs in a stereo manner and test with the resolution of $1024\times320$. 

In addition, to compare with Robust-Depth$^*$ \cite{iccv} fairly, we only adopt the WeatherDepth$^*$ in comparative experiments as shown in Table \ref{table2}, because both Robust-Depth$^*$ and WeatherDepth$^*$ share the same baseline (MonoViT),  training/testing resolutions and utilize the weather-augmented dataset for training. You can also check the comparing qualitative results in the supplement materials. As for WeatherDepth and WeatherDepth$^{\dagger}$, they adopt different resolutions as used in \cite{pld, wav} and different baselines. To prevent unfair comparison, we only report the performance gains with the proposed curriculum contrastive learning scheme, shown in Table \ref{table3} and Table \ref{table6}. 

\vspace{-0.3cm}
\begin{table}[ht]\centering\scriptsize
\caption{\footnotesize \centering  Model implementation details.}
 \vspace{-0.3cm}
\label{table1}
\scalebox{1}
{
\begin{tabular}{c|c|c|c}
\hline
 \textbf{Baseline}& \textbf{Name}& \textbf{Threshold} &  \textbf{$w_{cst}$}\\
\hline
PlaneDepth & WeatherDepth &0&0.01\\
MonoViT & WeatherDepth$^*$&5e-4&0.02\\
WaveletMonoDepth &WeatherDepth$^{\dagger}$&0&0.1\\
\hline
\end{tabular}
}
\vspace{-0.3cm}
\end{table}

\subsection{WeatherKITTI Results}
We show detailed comparative experiments between our method and current SoTA models in Table \ref{table2}(a). The Eigen raw split \cite{eigen} is used for evaluation following common practice, and we report the average results under 7 different weather conditions. 

In particular, Robust-Depth$^*$ \cite{iccv}is the latest SoTA model that tries to tackle weather conditions like us, but our WeatherDepth$^{*}$ greatly outperforms it with the same baseline (MonoViT). These results sufficiently demonstrate that our method is able to handle weather variations and domain changes.
As for the other two baselines, our WeatherDepth and WeatherDepth$^{\dagger}$ have also shown a significant gain in Table \ref{table3} (a) and Table \ref{table6} (a), which implies that our strategy can be generalized to other typical depth estimation methods.

\vspace{-0.2cm}
\begin{table}[ht]\centering\scriptsize
\caption{\footnotesize \centering Quantitative Comparison against SoTA Models}
\label{table2}
 \vspace{-0.2cm}
\scalebox{0.9}
{
\begin{tabular}{c|c|c|c|c|c|c|c}
\hline
 \textbf{Method} &  \textbf{absrel} &    \textbf{sqrel} &     \textbf{rmse} & \textbf{rmselog} &  \textbf{a1} &   \textbf{a2} &    \textbf{a3}  \\
 \hline
\multicolumn{8}{c} {\textbf{(a) WeatherKITTI}}\\
\hline
MonoViT\cite{vit}&   0.120  &   0.899  &   5.111  &   0.200  &   0.857  &   0.953  &   0.980  \\
PlaneDepth\cite{pld} &   0.150  &   1.360  &   6.513  &   0.277  &   0.757  &   0.891  &   0.945  \\
Robust-Depth$^*$\cite{iccv}&   0.107  &   0.791  &   4.604  &   0.183  &   0.883  &   0.963  &   0.983 \\
\textbf{WeatherDepth$^{*}$} &  \textbf{0.103} &  \textbf{0.738}  &  \textbf{4.414} &  \textbf{0.178}  &  \textbf{0.892} &  \textbf{0.965} &  \textbf{0.984}\\
\hline
\multicolumn{8}{c} {\textbf{(b) DrivingStereo Dataset:Rain}}\\
\hline
MonoViT\cite{vit}& 0.175 & 2.138 & 9.616 & 0.232 & 0.730 & 0.931 & 0.979 \\
PlaneDepth \cite{pld}& 0.220 & 3.302 & 11.671 & 0.278 & 0.654 & 0.883 & 0.965 \\
Robust-Depth$^*$\cite{iccv} & 0.167 & 2.019 & 9.157 & 0.221 & 0.755 & 0.938 & 0.982 \\
\textbf{WeatherDepth$^*$} & \textbf{0.158} & \textbf{1.833} & \textbf{8.837} & \textbf{0.211} & \textbf{0.764} & \textbf{0.945} & \textbf{0.985} \\
\hline
\multicolumn{8}{c} {\textbf{(c) CADC Dataset: Snow}}\\
\hline
MonoViT\cite{vit}& 0.297 & \textbf{4.499} & 10.757 & 0.369 & 0.547 & 0.835 & 0.935 \\
PlaneDepth\cite{pld}&   0.356  &   4.903  &  11.453  &   0.405  &   0.447  &   0.749  &   0.908  \\
Robust-Depth$^*$ \cite{iccv}& 0.298 & 5.550 & 11.481 & 0.369 & 0.590 & 0.853 & 0.932 \\
\textbf{WeatherDepth$^*$}& \textbf{0.279} & 4.632 & \textbf{10.699} & \textbf{0.357} & \textbf{0.597} &\textbf{0.857} & \textbf{0.938} \\
\hline
\multicolumn{8}{c} {\textbf{(d) DrivingStereo Dataset: Fog}}\\
\hline
MonoViT\cite{vit}& 0.109 & 1.206 & 7.758 & 0.167 & 0.870 & 0.967 & 0.990 \\
PlaneDepth\cite{pld}& 0.151 & 1.836 & 9.350 & 0.209 & 0.803 & 0.945 & 0.983 \\
Robust-Depth$^*$ \cite{iccv}& \textbf{0.105} & 1.135 & \textbf{7.276} &\textbf{0.158} & \textbf{0.882} & \textbf{0.974} & \textbf{0.992}\\
\textbf{WeatherDepth$^*$}  &\textbf{0.105}  &   \textbf{1.117}  &   7.346  &   \textbf{0.158}  &   0.879  &   0.972  &  \textbf{0.992} \\
\hline
\end{tabular}
}
\vspace{-0.1cm}
\end{table}

\vspace{-0.1cm}
\begin{table}[ht]\centering
\footnotesize 
\vspace{-0.3cm}
\caption{\footnotesize \centering  Incremental performance on PlaneDepth}
 \vspace{-0.2cm}
\scalebox{0.8}{
\begin{tabular}{c|c|c|c|c|c|c|c}
\hline
 \textbf{Method} &  \textbf{absrel} &    \textbf{sqrel} &     \textbf{rmse} & \textbf{rmselog} &  \textbf{a1} &   \textbf{a2} &    \textbf{a3}  \\
 \hline
\multicolumn{8}{c} {\textbf{(a) WeatherKITTI}}\\
\hline
 PlaneDepth\cite{pld}&   0.158  &   1.585  &   6.603  &   0.315  &   0.753  &   0.862  &   0.915  \\
% WeatherDepth w/o CC &   0.102  &   0.689  &   4.383  &   0.189  &   0.883  &   \textbf{0.960}  &   \textbf{0.981}  \\
 \textbf{WeatherDepth} &   \textbf{0.099}  &  \textbf{0.673} &   \textbf{4.324}  &   \textbf{0.185}  &   \textbf{0.884}  &  \textbf{0.959}  &   \textbf{0.981}  \\
\hline
\multicolumn{8}{c} {\textbf{(b) DrivingStereo Dataset:Rain}}\\
\hline
PlaneDepth\cite{pld}& 0.215 & 3.659 & 12.112 & 0.271 & 0.670 & 0.889 & 0.964 \\
% WeatherDepth w/o CC &   0.211  &   3.131  &  10.861  &   0.265  &   0.653  &   0.888  &   0.971  \\
\textbf{WeatherDepth} & \textbf{0.166} & \textbf{1.874} & \textbf{8.844} & \textbf{0.217} & \textbf{0.748} & \textbf{0.942} & \textbf{0.985} \\
\hline   
\multicolumn{8}{c} {\textbf{(c) CADC Dataset: Snow}}\\
\hline
PlaneDepth\cite{pld}& 0.367 & 5.509 & 11.845 & 0.420 & 0.436 & 0.743 & 0.897 \\
% WeatherDepth w/o CC &   0.287  &   4.871  &  11.322  &   0.362  &   0.582  &   0.845  &   0.933  \\
\textbf{WeatherDepth}& \textbf{0.278} & \textbf{4.220} & \textbf{10.571} & \textbf{0.353} & \textbf{0.585} & \textbf{0.854} & \textbf{0.937}\\
 \hline
\multicolumn{8}{c} {\textbf{(d)) DrivingStereo Dataset: Fog}}\\
\hline
PlaneDepth\cite{pld}&\textbf{0.122}& 1.416 & 8.306 & 0.179 & 0.847 & 0.961 & 0.990 \\
% WeatherDepth w/o CC &   0.131  &   1.551  &   7.935  &   0.180  &   0.843  &   0.964  &   0.990  \\
\textbf{WeatherDepth}& 0.123 & \textbf{1.404} & \textbf{7.679} &\textbf{0.172} &\textbf{0.859} &\textbf{0.968} &\textbf{0.992} \\
\hline   

\end{tabular}
}
\label{table3}
 \vspace{-0.5cm}
\end{table}

%%%%%wavlet
\begin{table}[ht]\centering\scriptsize
\caption{\footnotesize \centering Incremental performance on WaveletMonodepth}
\label{table6}
 \vspace{-0.2cm}
\scalebox{0.85}
{
\begin{tabular}{c|c|c|c|c|c|c|c}
\hline
 \textbf{Method} &  \textbf{absrel} &    \textbf{sqrel} &     \textbf{rmse} & \textbf{rmselog} &  \textbf{a1} &   \textbf{a2} &    \textbf{a3}  \\
 \hline
\multicolumn{8}{c} {\textbf{(a) WeatherKITTI}}\\
\hline
WaveletMonodepth\cite{wav}&   0.164  &   1.540  &   6.576  &   0.309  &   0.737  &   0.866  &   0.925 
% \\\\
% WeatherDepth$^{\dagger}$ w/o CC&   0.104  &   0.825  &   4.628  &   0.192  &  \textbf{0.878}  &   \textbf{0.958}  &   \textbf{0.981} \\
% WeatherDepth$^{\dagger}$  w/o C &   0.105  &   0.840  &   4.659  &   0.194  &   0.875  &   0.958  &   0.980  \\
\\
\textbf{WeatherDepth$^{\dagger}$}&  \textbf{0.103}  &  \textbf{0.777} &  \textbf{4.532} &  \textbf{0.191} &  \textbf{0.878}& \textbf{0.958}&  \textbf{0.981} \\
\hline
\multicolumn{8}{c} {\textbf{(b) DrivingStereo Dataset:Rain}}\\
\hline
WaveletMonodepth\cite{wav}& 0.280 & 4.604 & 13.231 & 0.339 & 0.570 & 0.819 & 0.922 \\
% WeatherDepth$^{\dagger}$ w/o CC &   0.220  &   3.206  &  11.291  &   0.279  &   0.655  &   0.880  &   0.958  \\
\textbf{WeatherDepth$^{\dagger}$} & \textbf{0.245} & \textbf{3.907} & \textbf{12.396} & \textbf{0.309} & \textbf{0.615} & \textbf{0.857} & \textbf{0.943} \\
\hline   
\multicolumn{8}{c} {\textbf{(c) CADC Dataset: Snow}}\\
\hline
WaveletMonodepth\cite{wav}& 0.503 & 8.361 & 14.529 & 0.514 & 0.314 & 0.591 & 0.798\\
% WeatherDepth$^{\dagger}$ w/o CC &   0.388  &   7.145  &  13.562  &   0.441  &   0.436  &   0.728  &   0.880  \\
\textbf{WeatherDepth$^{\dagger}$}& \textbf{0.394} & \textbf{6.828} & \textbf{13.293} & \textbf{0.443} & \textbf{0.427} & \textbf{0.714} & \textbf{0.876} \\
 \hline
\multicolumn{8}{c} {\textbf{(d)) DrivingStereo Dataset: Fog}}\\
\hline
WaveletMonodepth\cite{wav}& 0.144 & 1.886 & 9.720 & 0.218 & 0.801 & 0.937 & 0.975\\
% WeatherDepth$^{\dagger}$ w/o CC &   \textbf{0.140}  &   \textbf{1.755}  &   \textbf{8.831}  &   \textbf{0.199}  &   \textbf{0.820}  &   \textbf{0.951}  &   0.983  \\
\textbf{WeatherDepth$^{\dagger}$}& \textbf{0.140} & \textbf{1.784} & \textbf{8.893} & \textbf{0.199} & \textbf{0.818} & \textbf{0.950} & \textbf{0.984} \\
\hline
\end{tabular}
}
 \vspace{-0.7cm}
\end{table}

\subsection{Real Weather Scenes Results}
To validate the ability to handle real-world weather, we follow WeatherKITTI and use real-world rainy, snowy, and foggy weather data for testing. 

Note that all of our models are only trained on WeatherKITTI (zero-shot for real-world datasets).

\subsubsection{Rain}
We show the quantitative results on real rainy data from the DrivingStereo dataset in Table \ref{table2} (b). Our method is still more accurate than the existing solutions, which demonstrates our model can adapt to the variations of rainy scenes. Especially, our scheme reduces the errors from water reflections and lens droplets as shown in Fig. \ref{Fig1}(c).

\subsubsection{Snow}
As depicted in Table \ref{table2} (c), our method reaches the SoTA performance on the new CADC dataset. 
This further suggests that our method enables the depth estimation models to ignore the erroneous depth cues (Fig. \ref{Fig1}(a)(b)) due to the progressive introduction of ground snow and snowflakes, which is very challenging for depth estimation tasks.

\subsubsection{Fog}
In Table \ref{table2} (d), we collect the results of fog scene evaluation for our model and SoTA frameworks. Unfortunately, the fog density in the DrivingStereo dataset is relatively light, while our model aims to address more 
adverse weather conditions (using more severe fog augmentation). Hence, we take the second-stage model here and achieve comparable results with Robust-Depth$^*$. 

In the (b-d) of Table \ref{table3} and Table \ref{table6}, we can notice a similar trend as that of Table \ref{table2}, which further demonstrates the generalization capacity of the proposed strategy on different models(WaveletMonodepth, MonoViT, and PlaneDepth) and on different real weather datasets(Rain, Snow, and Fog).

\vspace{-0.1cm}
\subsection{Ablation Study}
We have demonstrated the superiority of our method on synthetic and real adverse weather data. Next, we conduct experiments to validate the effectiveness of each component. For clarity, we only take WeatherDepth$^*$ in the ablation study. As for the other models, they show similar performance and will be demonstrated in the supplement materials. 

\subsubsection{Learning Strategy}
We define the direct mixed training manner as "w/o CC", in which each weather condition has $1/n$ probability of being selected for training. As shown in Tables \ref{table4}(a-e), since mixed training attempts to estimate depth from erroneous depth cues introduced by weather variation, this strategy performs poorly on our WeatherKITTI dataset. Moreover, it degrades the performance across all three real weather scenes, further validating the effectiveness of our proposed curriculum contrastive learning.

\begin{table}[ht]\centering
\footnotesize
\caption{\footnotesize \centering  Quantitative Results on Ablation Study}
\label{table4}
 \vspace{-0.25cm}
\scalebox{0.75}{
\begin{tabular}{c|c|c|c|c|c|c|c}
\hline
 \textbf{Method} &  \textbf{absrel} &    \textbf{sqrel} &     \textbf{rmse} & \textbf{rmselog} &  \textbf{a1} &   \textbf{a2} &    \textbf{a3}  \\
\hline
\multicolumn{8}{c} {\textbf{(a) WeatherKITTI}}\\
\hline
MonoViT\cite{vit}&   0.120  &   0.899  &   5.111  &   0.200  &   0.857  &   0.953  &   0.980  \\
WeatherDepth$^*$ w/o CC&   0.105  &   0.833  &   4.554  &   0.180  &   \textbf{0.893}  &   0.964  &   0.983  \\
WeatherDepth$^*$ w/o C&   0.104  &   0.772  &   4.527  &   0.181  &   0.890  &   0.964  &   0.983  \\
\textbf{ WeatherDepth$^{*}$} &  \textbf{0.103} &  \textbf{0.738}  &  \textbf{4.414} &  \textbf{0.178}  &  0.892 &  \textbf{0.965} &  \textbf{0.984}\\
 \hline
\multicolumn{8}{c} {\textbf{(b) DrivingStereo Dataset:Rain}}\\
\hline
MonoViT\cite{vit}& 0.175 & 2.138 & 9.616 & 0.232 & 0.730 & 0.931 & 0.979 \\
WeatherDepth$^*$ w/o CC&   0.163  &   1.949  &   9.124  &   0.216  &   0.759  &   0.944  &   0.984  \\
WeatherDepth$^*$ w/o C&   \textbf{0.156}  &   \textbf{1.755}  &   8.916  &   0.212  &   \textbf{0.768}  &   \textbf{0.945}  &   \textbf{0.985}  \\
\textbf{WeatherDepth$^*$} & 0.158 & 1.833 & \textbf{8.837} & \textbf{0.211} & 0.764 & \textbf{0.945} & \textbf{0.985} \\
\hline   
\multicolumn{8}{c} {\textbf{(c) CADC Dataset: Snow}}\\
\hline
MonoViT\cite{vit}& 0.297 & \textbf{4.499} & 10.757 & 0.369 & 0.547 & 0.835 & 0.935 \\
WeatherDepth$^*$ w/o CC&   0.305  &   5.967  &  11.846  &   0.373  &   0.582  &   0.847  &   0.929  \\
WeatherDepth$^*$ w/o C &   0.306  &   5.476  &  11.261  &   0.371  &   0.562  &   0.843  &   0.934  \\
\textbf{WeatherDepth$^*$}&\textbf{0.279} & 4.632 & \textbf{10.699} & \textbf{0.357} & \textbf{0.597} &\textbf{0.857} & \textbf{0.938}\\
 \hline
\multicolumn{8}{c} {\textbf{(d)) DrivingStereo Dataset: Fog}}\\
\hline
MonoViT\cite{vit}& 0.109 & 1.206 & 7.758 & 0.167 & 0.870 & 0.967 & 0.990 \\
WeatherDepth$^*$ w/o CC &   0.112  &   1.282  &   7.511  &   0.163  &   0.873  &   0.970  &   \textbf{0.992}  \\
WeatherDepth$^*$ w/o C&   \textbf{0.106}  &   \textbf{1.145}  &   7.434  &   0.161  &   \textbf{0.880}  &   0.971  &   0.991  \\
\textbf{WeatherDepth$^*$}  & 0.110 & 1.195 & \textbf{7.323} & \textbf{0.160} & 0.878 & \textbf{0.973} & \textbf{0.992} \\
\hline   
\multicolumn{8}{c} {\textbf{(e) KITTI (only sunny scenes)}}\\
\hline
MonoViT\cite{vit} & \textbf{0.099}&	0.708	&4.372	&0.175	&\textbf{0.900}	&\textbf{0.967}	&\textbf{0.984}\\
WeatherDepth$^*$ w/o CC&   0.104  &   0.826  &   4.532  &   0.179  &   0.896  &   0.965  &   0.983  \\
WeatherDepth$^*$ w/o C&   0.100  &   0.702  &   4.445  &   0.177  &   0.894  &   0.965  &   \textbf{0.984}  \\
\textbf{WeatherDepth$^*$}&  \textbf{0.099}&   \textbf{0.698} &  \textbf{4.330}  &   \textbf{0.174} &   0.897  &   \textbf{0.967} &   \textbf{0.984}  \\
 \hline
\end{tabular}
}
\vspace{-0.6cm}
\end{table}

\subsubsection{Robustness on Sunny Condition}
As reported in Table \ref{table4}(a-d), WeatherDepth$^*$ without contrastive learning integration (termed as "w/o C") shows a competitive performance with our final model on real and synthesized weather data. However, its performance on sunny days sharply declines as illustrated in Table \ref{table4}(e). In other words,  without contrastive learning, the network only carries out weather domain transferring and loses the previous knowledge to tackle sunny scenes. 
With our curriculum contrastive learning, our performance(trained on WeatherKITTI) on sunny data even outperforms MonoViT\cite{vit}(that trained on sunny data), which further suggests that our training manner gives robust depth results in any conditions.

\subsubsection{Efficient Contrastive Learning}
We incorporate the contrastive strategy of Robust-Depth and our contrastive scheme into the ResNet-18-based WaveletMonodepth model and train for 30 epochs on a 2080ti GPU. As shown in Table \ref{table5}, our training time increases by less than 20\% compared to no contrastive learning, while the Robust-Depth nearly doubles the training time due to bilateral propagation and image reconstruction. This validates that our method addresses performance degradation under weather conditions in a more efficient mode.

\begin{table}[ht]\centering
\footnotesize
\vspace{-0.35cm}
\caption{\footnotesize \centering  Verify the efficiency of our Contrast Learning Manner}
\label{table5}
\scalebox{0.9}{
\begin{tabular}{c|c|c|c}
\hline
 \textbf{Contrast Method} &  Robust-Depth Way&WeatherDepth Way& No Contrast \\
  \hline
 \textbf{Training Time}&20h02m41s&12h10m07s&10h39m25s\\
 \hline
\end{tabular}
}
\vspace{-0.5cm}
\end{table}

\vspace{-0.15cm}
\section{Conclusion}
\vspace{-0.1cm}
\label{conclusion}
In this paper, we propose WeatherDepth, a self-supervised robust depth estimation strategy that employs curriculum contrastive learning to effectively tackle performance degradation in complex weather conditions. Our model progressively adapts to adverse weather scenes through a curriculum learning scheme and is further combined with contrastive learning to prevent knowledge forgetfulness. To narrow the weather domain gap, we also build the WeatherKITTI dataset to help models better adapt to real weather scenarios. Through extensive experiments, we demonstrate the effectiveness of our WeatherDepth against various architectures and its superior performance over existing SoTA solutions.

\vspace{-0.15cm}
\section{Acknowledgement}
\vspace{-0.1cm}
This work was supported by the National Natural Science Foundation of China (Nos. 62172032,62372036).

\normalem
\bibliographystyle{plain}
\bibliography{ref}

\begin{thebibliography}{10}

\bibitem{r19}
Juan Luis~Gonzalez Bello and Munchurl Kim.
\newblock Self-supervised deep monocular depth estimation with ambiguity
  boosting.
\newblock {\em IEEE Transactions on Pattern Analysis and Machine Intelligence},
  44(12):9131--9149, 2022.

\bibitem{r24}
Yoshua Bengio, J{\'e}r{\^o}me Louradour, Ronan Collobert, and Jason Weston.
\newblock Curriculum learning.
\newblock In {\em International Conference on Machine Learning}, 2009.

\bibitem{nuscenes}
Holger Caesar, Varun Bankiti, Alex~H. Lang, Sourabh Vora, Venice~Erin Liong,
  Qiang Xu, Anush Krishnan, Yu~Pan, Giancarlo Baldan, and Oscar Beijbom.
\newblock nuscenes: A multimodal dataset for autonomous driving, 2020.

\bibitem{dror}
Nicholas Charron, Stephen Phillips, and Steven Waslander.
\newblock De-noising of lidar point clouds corrupted by snowfall.
\newblock pages 254--261, 05 2018.

\bibitem{r27}
Wei-Ting Chen, Hao-Yu Fang, Cheng-Lin Hsieh, Cheng-Che Tsai, I~Chen, Jian-Jiun
  Ding, Sy-Yen Kuo, et~al.
\newblock All snow removed: Single image desnowing algorithm using hierarchical
  dual-tree complex wavelet representation and contradict channel loss.
\newblock In {\em Proceedings of the IEEE/CVF International Conference on
  Computer Vision}, pages 4196--4205, 2021.

\bibitem{eigen}
David Eigen, Christian Puhrsch, and Rob Fergus.
\newblock Depth map prediction from a single image using a multi-scale deep
  network, 2014.

\bibitem{r17}
Ravi Garg, Vijay~Kumar BG, Gustavo Carneiro, and Ian Reid.
\newblock Unsupervised cnn for single view depth estimation: Geometry to the
  rescue, 2016.

\bibitem{iccv2}
Stefano Gasperini, Nils Morbitzer, HyunJun Jung, Nassir Navab, and Federico
  Tombari.
\newblock Robust monocular depth estimation under challenging conditions.
\newblock In {\em Proceedings of the IEEE/CVF International Conference on
  Computer Vision}, 2023.

\bibitem{kitti}
Andreas Geiger, Philip Lenz, and Raquel Urtasun.
\newblock Are we ready for autonomous driving? the kitti vision benchmark
  suite.
\newblock In {\em 2012 IEEE conference on computer vision and pattern
  recognition}, pages 3354--3361. IEEE, 2012.

\bibitem{mdp1}
Clément Godard, Oisin~Mac Aodha, and Gabriel~J. Brostow.
\newblock Unsupervised monocular depth estimation with left-right consistency,
  2017.

\bibitem{r13}
Clément Godard, Oisin~Mac Aodha, Michael Firman, and Gabriel Brostow.
\newblock Digging into self-supervised monocular depth estimation, 2019.

\bibitem{r1}
Shahram Izadi, David Kim, Otmar Hilliges, David Molyneaux, Richard Newcombe,
  Pushmeet Kohli, Jamie Shotton, Steve Hodges, Dustin Freeman, Andrew Davison,
  et~al.
\newblock Kinectfusion: real-time 3d reconstruction and interaction using a
  moving depth camera.
\newblock In {\em Proceedings of the 24th annual ACM symposium on User
  interface software and technology}, pages 559--568, 2011.

\bibitem{rbt}
Lingdong Kong, Yaru Niu, Shaoyuan Xie, Hanjiang Hu, Lai~Xing Ng, Benoit~R.
  Cottereau, Ding Zhao, Liangjun Zhang, Hesheng Wang, Wei~Tsang Ooi, Ruijie
  Zhu, Ziyang Song, Li~Liu, Tianzhu Zhang, Jun Yu, Mohan Jing, Pengwei Li,
  Xiaohua Qi, Cheng Jin, Yingfeng Chen, Jie Hou, Jie Zhang, Zhen Kan, Qiang
  Ling, Liang Peng, Minglei Li, Di~Xu, Changpeng Yang, Yuanqi Yao, Gang Wu,
  Jian Kuai, Xianming Liu, Junjun Jiang, Jiamian Huang, Baojun Li, Jiale Chen,
  Shuang Zhang, Sun Ao, Zhenyu Li, Runze Chen, Haiyong Luo, Fang Zhao, and
  Jingze Yu.
\newblock The robodepth challenge: Methods and advancements towards robust
  depth estimation, 2023.

\bibitem{r22}
Younkwan Lee, Jihyo Jeon, Yeongmin Ko, Byunggwan Jeon, and Moongu Jeon.
\newblock Task-driven deep image enhancement network for autonomous driving in
  bad weather, 2021.

\bibitem{r5}
Lina Liu, Xibin Song, Mengmeng Wang, Yong Liu, and Liangjun Zhang.
\newblock Self-supervised monocular depth estimation for all day images using
  domain separation, 2021.

\bibitem{r14}
Rui Peng, Ronggang Wang, Yawen Lai, Luyang Tang, and Yangang Cai.
\newblock Excavating the potential capacity of self-supervised monocular depth
  estimation, 2021.

\bibitem{cadc}
Matthew Pitropov, Danson~Evan Garcia, Jason Rebello, Michael Smart, Carlos
  Wang, Krzysztof Czarnecki, and Steven Waslander.
\newblock Canadian adverse driving conditions dataset.
\newblock {\em The International Journal of Robotics Research},
  40(4-5):681--690, 2021.

\bibitem{r28}
Lutz Prechelt.
\newblock Early stopping-but when?
\newblock In {\em Neural Networks: Tricks of the trade}, pages 55--69.
  Springer, 2002.

\bibitem{wav}
Micha{\"{e}}l Ramamonjisoa, Michael Firman, Jamie Watson, Vincent Lepetit, and
  Daniyar Turmukhambetov.
\newblock Single image depth prediction with wavelet decomposition.
\newblock In {\em Proceedings of the IEEE/CVF Conference on Computer Vision and
  Pattern Recognition}, June 2021.

\bibitem{acdc}
Christos Sakaridis, Dengxin Dai, and Luc Van~Gool.
\newblock Acdc: The adverse conditions dataset with correspondences for
  semantic driving scene understanding.
\newblock In {\em Proceedings of the IEEE/CVF International Conference on
  Computer Vision}, pages 10765--10775, 2021.

\bibitem{iccv}
Kieran Saunders, George Vogiatzis, and Luis Manso.
\newblock Self-supervised monocular depth estimation: Let's talk about the
  weather, 2023.

\bibitem{r8}
Maxime Tremblay, Shirsendu~Sukanta Halder, Raoul de~Charette, and
  Jean-Fran{\c{c}}ois Lalonde.
\newblock Rain rendering for evaluating and improving robustness to bad
  weather.
\newblock {\em International Journal of Computer Vision}, 129(2):341--360, sep
  2020.

\bibitem{pld}
Ruoyu Wang, Zehao Yu, and Shenghua Gao.
\newblock Planedepth: Self-supervised depth estimation via orthogonal planes,
  2023.

\bibitem{r9}
Xin Wang, Yudong Chen, and Wenwu Zhu.
\newblock A survey on curriculum learning.
\newblock {\em IEEE Transactions on Pattern Analysis and Machine Intelligence},
  44(9):4555--4576, 2022.

\bibitem{r15}
Jamie Watson, Oisin~Mac Aodha, Victor Prisacariu, Gabriel Brostow, and Michael
  Firman.
\newblock {The Temporal Opportunist: Self-Supervised Multi-Frame Monocular
  Depth}.
\newblock In {\em Computer Vision and Pattern Recognition (CVPR)}, 2021.

\bibitem{stereo}
Guorun Yang, Xiao Song, Chaoqin Huang, Zhidong Deng, Jianping Shi, and Bolei
  Zhou.
\newblock Drivingstereo: A large-scale dataset for stereo matching in
  autonomous driving scenarios.
\newblock In {\em IEEE Conference on Computer Vision and Pattern Recognition
  (CVPR)}, 2019.

\bibitem{r20}
Zehao Yu, Lei Jin, and Shenghua Gao.
\newblock P 2 net: Patch-match and plane-regularization for unsupervised indoor
  depth estimation.
\newblock In {\em European Conference on Computer Vision}, pages 206--222.
  Springer, 2020.

\bibitem{r2}
Yuxiao Zhang, Alexander Carballo, Hanting Yang, and Kazuya Takeda.
\newblock Perception and sensing for autonomous vehicles under adverse weather
  conditions: A survey.
\newblock {\em {ISPRS} Journal of Photogrammetry and Remote Sensing},
  196:146--177, feb 2023.

\bibitem{r3}
Chaoqiang Zhao, Yang Tang, and Qiyu Sun.
\newblock Unsupervised monocular depth estimation in highly complex
  environments, 2022.

\bibitem{vit}
Chaoqiang Zhao, Youmin Zhang, Matteo Poggi, Fabio Tosi, Xianda Guo, Zheng Zhu,
  Guan Huang, Yang Tang, and Stefano Mattoccia.
\newblock {MonoViT}: Self-supervised monocular depth estimation with a vision
  transformer.
\newblock In {\em 2022 International Conference on 3D Vision (3DV)}. {IEEE},
  sep 2022.

\bibitem{zhou}
Tinghui Zhou, Matthew Brown, Noah Snavely, and David~G. Lowe.
\newblock Unsupervised learning of depth and ego-motion from video, 2017.

\bibitem{cyclegan}
Jun-Yan Zhu, Taesung Park, Phillip Isola, and Alexei~A Efros.
\newblock Unpaired image-to-image translation using cycle-consistent
  adversarial networks.
\newblock In {\em Proceedings of the IEEE international conference on computer
  vision}, pages 2223--2232, 2017.

\end{thebibliography}
\end{document}